\documentclass[letterpaper]{article} 
\usepackage{aaai25}  
\usepackage{times}  
\usepackage{helvet}  
\usepackage{courier}  
\usepackage[hyphens]{url}  
\usepackage{graphicx} 
\usepackage{natbib}  
\usepackage{caption} 
\usepackage{algorithm}
\usepackage{algorithmic}
\usepackage{amsmath}
\usepackage{booktabs}
\usepackage{multirow}
\usepackage{newfloat}
\usepackage{listings}
\DeclareCaptionStyle{ruled}{labelfont=normalfont,labelsep=colon,strut=off} 
\floatstyle{ruled}
\newfloat{listing}{tb}{lst}{}
\floatname{listing}{Listing}
\title{An Efficient Replay for Class-Incremental Learning with Pre-trained Models}
\author{
    Weimin Yin,
    Bin Chen,
    Chunzhao Xie,
    Zhenhao Tan
}
\affiliations{

}

\usepackage{bibentry}

\begin{document}

\maketitle

\begin{abstract}
In general class-incremental learning, researchers typically use sample sets as a tool to avoid catastrophic forgetting during continuous learning. At the same time, researchers have also noted the differences between class-incremental learning and Oracle training and have attempted to make corrections. In recent years, researchers have begun to develop class-incremental learning algorithms utilizing pre-trained models, achieving significant results. This paper observes that in class-incremental learning,  the steady state among the weight guided by each class center is disrupted, which is significantly correlated with catastrophic forgetting. Based on this, we propose a new method to overcoming forgetting . In some cases, by retaining only a single sample unit of each class in memory for replay and applying simple gradient constraints, very good results can be achieved. Experimental results indicate that under the condition of pre-trained models, our method can achieve competitive performance with very low computational cost and by simply using the cross-entropy loss.

\end{abstract}

%

\section{Introduction}

In the real world, data is continuously and dynamically generated, while the current mainstream training methods require pre-collecting large amounts of data. However, when a model is trained on non-stationary data, with different classification tasks arriving sequentially, it leads to catastrophic forgetting 
 \cite{mccloskey1989catastrophic}, resulting in a decline in performance on previously learned data. This undoubtedly limits the flexibility of current artificial intelligence to some extent.

Research on catastrophic forgetting is widely distributed in the fields of continual learning or incremental learning. Recently, pre-trained models have been shown to be beneficial for continual learning, and many methods have achieved significant results. However, to further study catastrophic forgetting, we need to temporarily return to the perspective of classical research. 
To overcome catastrophic forgetting, extensive research has focused on continuously adjusting the entire or partial weights of the model as the data distribution changes, aiming to retain knowledge from previous tasks \cite{zhou2023deep,de2021continual}. This typically relies on a sample buffer to retrain portions of past samples, and this method is believed to be rooted in the complementary learning systems theory \cite{kumaran2016learning}, inspired by hippocampal episodic memory. On the other hand, since catastrophic forgetting is a phenomenon arising from atypical training processes, it can be viewed as the abnormal variation of the forgetting model's parameters relative to normal training. This branch of research \cite{castro2018end,wu2019large,hou2019learning} seeks to rectify forgetting issues by detecting differences between normal training and the training process that leads to forgetting, with the main goal being to align the form of the forgetting model with the normal model during training. Nevertheless, the primary defense against forgetting remains the use of a sample buffer.

\begin{figure}[t]
\centering
\includegraphics[width=1.0\columnwidth]{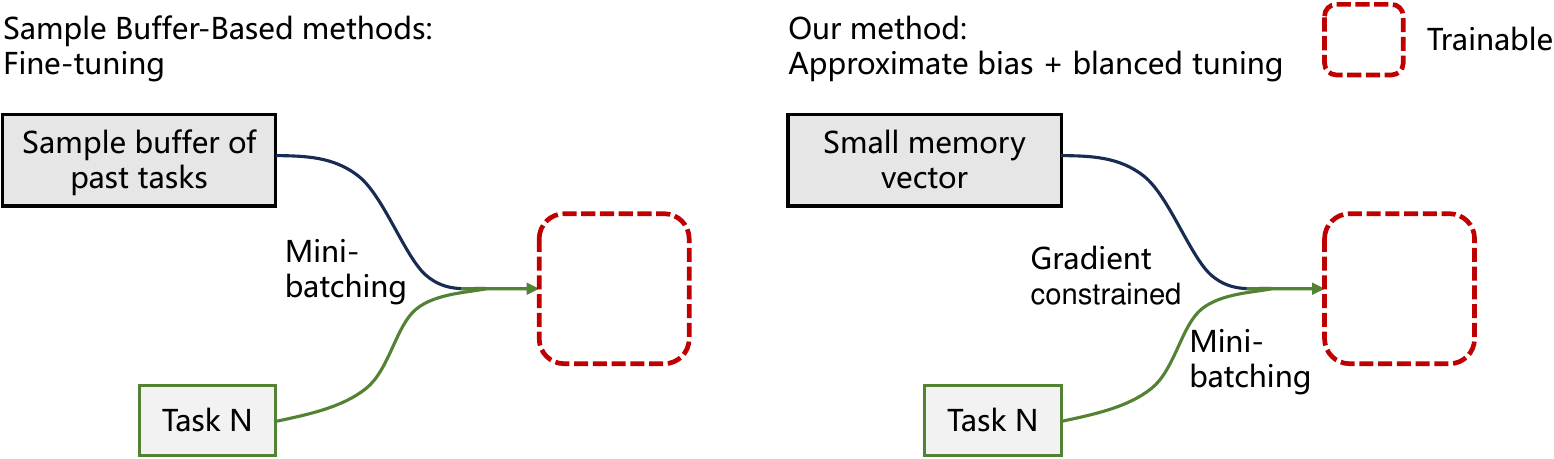} 
\caption{Overview of the WBR Framework. In contrast to typical methods that use sample buffers to incrementally adjust the entire or partial model weights to avoid catastrophic forgetting, WBR utilizes a single memory vector to represent all samples within a task, guiding the model to prevent forgetting. During the supervised learning of new tasks, WBR balances the proportion of old and new tasks in the model's weights by controlling the magnitude of gradient updates. Experimental results demonstrate that this balance is directly related to the occurrence of catastrophic forgetting. Notably, the maximum memory pool size we use is smaller than that of a single 224x224 image.}
\label{fig1}
\end{figure}

Based on previous research, we raise the following question: If the use of a sample buffer is merely to make the forgetting model closer to the normally trained model after training, can we, through another method, quickly eliminate this difference once the parameter differences between the forgetting model and the normal model are identified, without the need to retain a large number of historical samples?

To address this question, we draw inspiration from recent research on model rectify in the field of continuous learning. Model rectify techniques focus on the imbalance between the weights of new tasks and old tasks in the classification layer caused by catastrophic forgetting. Based on this difference, they design specific normalization or training methods to make the classification layer more consistent with the form of the normal model during continual learning \cite{zhao2020maintaining}. Intuitively, model rectify methods, in addition to using sample buffers to review past knowledge, add corrections to the classification layer. These corrections further mitigate forgetting in the classification layer. We believe it is of great significance to identify the patterns of differences between the forgetting model and the normal model.

Nevertheless, how to quickly fit the parameters of the forgetting model to the normal model to solve the aforementioned problem remains unclear. On one hand, like many algorithms, model rectify-based algorithms still require the help of a sample buffer to combat forgetting. On the other hand, as a global phenomenon, catastrophic forgetting involves all parameters, not just the classification layer. Therefore, if we only focus on the classification layer, it may be difficult to propose a comprehensive framework. This would hinder the understanding of the catastrophic forgetting.

To this end, starting from the same conceptual basis as previous ideas, we further extend this to arbitrary weight changes due to forgetting. To validate this concept, we propose Weight Balancing Replay (WBR). Figure \ref{fig1} provides an overview comparing our method with the sample buffer-based methods. WBR maximizes the utilization of the imbalance between the weights of new and old tasks in the forgetting model. During continual learning, for each past task, it retains only one vector representing the memory of that task, dynamically maintaining the weight balance while training new tasks. Specifically, these memory vectors are sampled from all samples within a task, similar to class centers. We designed a balancing mechanism using two hyperparameters to explicitly control the gradient update of new task data and memory vectors during training. The weight balancing constraint the optimization, allowing new knowledge and old knowledge to integrate rather than experiencing catastrophic forgetting. Our design explicitly targets the mechanism of catastrophic forgetting, thereby combating forgetting in the shortest path during the optimization process. Strictly speaking, WBR does not belong to the typical continuous learning training paradigm; it is more akin to a rapid simulation of a complementary learning system. Learned knowledge is stored in a compressed form, and when learning new knowledge, it integrates information by recalling the memory (Figure 2). Without a large number of samples in a sample buffer and the additional computational overhead, WBR has very high training efficiency.

In summary, this work has the following contributions:
\begin{itemize}
\item We propose Weight Balancing Replay (WBR), a new class-incremental learning method based on weight balancing. By controlling the gradient update size, WBR balances the changes in weights between new and old tasks during training, using a memory vector of a single sample size instead of a sample buffer. This method is particularly suitable for sequential learning tasks on pre-trained models and can significantly reduce the high training costs associated with large-scale models.
\item We conducted multiple experiments under class-incremental learning settings to demonstrate the effectiveness of WBR. In a toy environment, we verified the high correlation between catastrophic forgetting and weight balancing. Surprisingly, on the latest pre-trained model-based continual learning benchmarks, WBR achieved competitive results with a simple and low-cost advantage, making it ideal for real-world scenarios with complex device environments.
\item To the best of our knowledge, we are the first to correct catastrophic forgetting on a microscopic scale. We hope our approach provides a different perspective on addressing the frontier challenges of catastrophic forgetting.

\end{itemize}

\section{Related Work}

Here, we compare our method with related work and discuss their differences. The approaches to solving or circumventing the issue of catastrophic forgetting can be divided into four main categories:

\textbf{Parameter Regularization-Based:}These methods attempt to measure the impact of each parameter on the network's importance and protect acquired knowledge by preserving the invariance of critical parameters \cite{kirkpatrick2017overcoming,chaudhry2018riemannian,zenke2017continual}. Although these methods address catastrophic forgetting by estimating and calculating parameter importance without using a sample buffer, they fail to achieve satisfactory performance under more challenging conditions \cite{rebuffi2017icarl}. In contrast, our method directly controls parameter updates to mitigate the effects of catastrophic forgetting, rather than achieving this indirectly through regularization penalties.

\textbf{Sample Buffer-Based:}These methods build a sample buffer to store samples from old tasks \cite{chaudhry2018efficient}, or use generative networks to generate samples \cite{shin2017continual}, for training alongside data from new tasks. Additionally, many methods have improved on this simple and effective idea by using techniques such as knowledge distillation \cite{buzzega2020dark,rebuffi2017icarl} and model rectify methods \cite{zhao2020maintaining,hou2019learning}. Sample buffer-based methods have achieved leading performance in various benchmarks in the past \cite{mai2022online}. However, the performance of these methods generally decreases as the buffer size reduces \cite{cha2021co2l}, and they are severely limited when considering security and privacy concerns \cite{chamikara2018efficient}. In contrast to store historical data, our method compresses past knowledge into minimal memory vectors, which guide the balanced fine-tuning of weights when training new tasks. This balance greatly impacts the occurrence of forgetting. Our method requires minimal historical information to significantly mitigate catastrophic forgetting, suggesting that there may still be room for optimization in sample buffer-based methods.

\textbf{Architecture-Based:}These methods aim to introduce an additional component to store knowledge when encountering a new task. This component can be realized by copying and expanding the network \cite{yan2021dynamically,wang2022foster,zhou2022model}, or by dividing the network into more sub-networks \cite{ke2020continual}. However, these methods require a large number of additional parameters, with early methods even saving a backbone network for each task. To achieve scalability under limited memory budgets, some improved methods still use sample buffers (e.g., knowledge distillation) to reduce redundancy. In contrast, WBR does not add any extra parameters and is fundamentally different from architecture-based methods in concept: WBR focuses on integrating new and old knowledge through fine-tuning weights without changing the network architecture. Most architecture-based methods aim to provide additional parameters to preserve knowledge. Additionally, there is a branch based on meta-learning \cite{beaulieu2020learning}, which attempts to enable the overall architecture to overcome forgetting by providing additional meta-learning components through meta-training. This is conceptually similar to our method, except that we provide prior knowledge, whereas they hope to obtain this knowledge through meta-learning.

\textbf{Pretrained Models-Based:}These methods leverage the powerful representation capabilities of pretrained models for continual learning. Prompt-based methods establish connections between pretrained knowledge and continual learning tasks by designing prompt pools, selecting prompts \cite{wang2022learning}, and combining prompts \cite{smith2023coda}. In addition, some studies extend prompt selection methods by incorporating multimodal information. Instead of manual selection, some studies utilize multimodal information to select appropriate prompts \cite{razdaibiedina2023progressive}. Model-mixing methods aim to create a set of models during continual learning and perform model ensemble \cite{wang2023isolation} or model merging \cite{gao2023unified} during inference. This type of method combines the strengths of multiple models, reducing the problem of catastrophic forgetting. However, creating and maintaining multiple models increases computational and storage costs, especially in resource-limited environments. Representation-based methods seek to exploit pretrained features for continual learning, bearing some relation to our approach. By adjusting the embedding function with a small learning rate and the classifier with a large learning rate, features can gradually fit while the classifier quickly adapts. After training, modeling and replaying class feature distributions can calibrate the classifier to resist forgetting \cite{zhang2023slca}. Other studies achieve state-of-the-art performance by efficiently tuning the pretrained model with additional parameter-efficient modules \cite{jia2022visual,chen2022adaptformer} or connecting feature representations of multiple task-specific backbone networks \cite{zhou2024expandable}. Unlike these methods, our approach directly adjusts classifier weights during continual learning training without post-processing step, significantly simplifying the process.

\section{Prerequisites}

\subsection{Continual learning protocols}

Continual learning is typically defined as training a machine learning model on non-stationary data from a sequence of tasks. Consider a series of $B$ training task datasets $\{D_{1},D_{2},\dots,D_{B}\}$, where $D_{b}=\{(x_{i}^{b},y_{i}^{b})\}_{i=1}^{n_{b}}$ represents the $b$-th task in the sequence, containing $n_{b}$ training instances. Each instance $x_{i}^{b} $ belongs to a label $y_{i}\in Y_{b}$, with $Y_{b}$ being the label space for task $b$. For any $b\ne b^{'}$, $Y_{b}\cap Y_{b^{'}}=\emptyset $. The objective is to train a single model $f(x;\theta):X\to Y $, parameterized by $\theta$, to predict the label $y=f(x;\theta)\in Y$ for any sample $x$ from any previously learned task. During the training of task $b$, the model may only have access to the data in $D_{b}$. 

Depending on the environmental settings, common continual learning scenarios are divided into task-incremental, class-incremental, and domain-incremental learning. Task-incremental learning assumes that the task identity is known during testing. Domain-incremental learning maintains the same set of classes for each task, only changing the distribution of $x$ between tasks. The goal of class-incremental learning is to continuously build a classifier that covers all classes encountered across tasks. This means that the model needs to learn new knowledge from task $b$ while retaining knowledge from previous tasks. Our paper addresses the more challenging class-incremental learning setting.

\subsection{Benchmark for Continual Learning with Pretrained Models}

\begin{figure*}[t]
\centering
\includegraphics[width=0.8\textwidth]{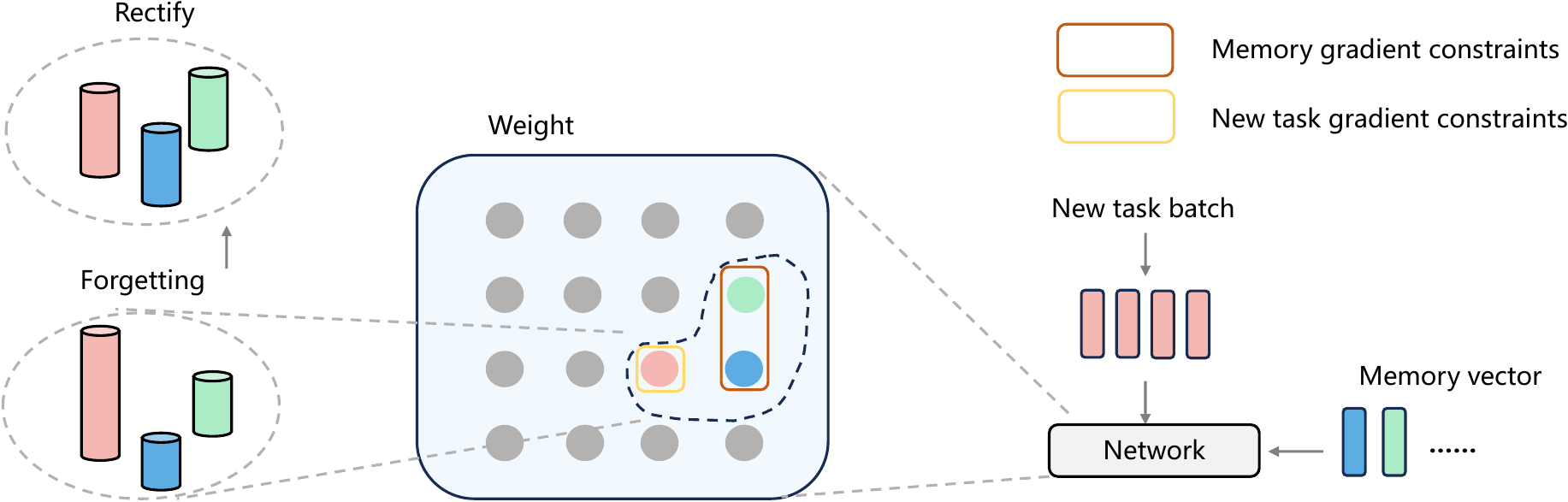} 
\caption{Illustration of WBR During Training. First, WBR samples and retains memory vectors during the learning of historical tasks based on our proposed bias approximation mechanism. Then, WBR incorporates all historical memory vectors into the training of the new task and optimizes the network using the loss function defined in Equation \ref{loss}. The goal is to learn the new task while maintaining a balance in the weights, guiding the network to avoid forgetting.}
\label{fig2}
\end{figure*}

Previous research typically trained a continual learning model from scratch. Recently, researchers have focused on designing continual learning algorithms based on pre-trained models (PTMs). Many studies have demonstrated that PTM-based continual learning methods can achieve excellent performance without the need for large sample buffers, thanks to the strong generalization capabilities of PTMs. From the perspective of representation learning, the essence of model training is to learn appropriate representations for each task, and a powerful PTM accomplishes this work from the very beginning. Therefore, the focus of continual learning has shifted to overcoming catastrophic forgetting. For example, techniques such as visual prompt tuning \cite{jia2022visual} and adapter learning \cite{chen2022adaptformer} can quickly adapt PTMs to downstream tasks while maintaining generalization. Consequently, compared to training from scratch, using PTMs in continual learning shows stronger performance in resisting forgetting \cite{cao2023retentive}.

Given the powerful representation capabilities of PTMs, if different PTMs lead to varying performance outcomes, how can we determine whether the differences are due to the algorithm or the PTM itself? In other words, how can we measure the fundamental continual learning ability provided by a PTM within an algorithm? SimpleCIL \cite{zhou2024revisitingclassincrementallearningpretrained}proposes a straightforward approach to achieve this. When faced with a continuous data stream, it freezes the pre-trained weights and extracts the center $c_{i}$ for each class $i$:
\begin{equation}
\label{equ_simplecil}
    c_{i}=\frac{1}{K} {\textstyle \sum_{j=1}^{|D_{b}|} I(y_{j}=i)\phi (x_{j}),  } 
\end{equation}
where $K={\textstyle \sum_{j=1}^{|D^{b}|} I(y_{j}=i) }$, The function $I(\cdot)$ returns 1 when the condition inside the parentheses is true and 0 when it is false. In equation \ref{equ_simplecil}, embeddings of the same class are averaged, leading to the most common pattern for the corresponding class. Accordingly, SimpleCIL directly replaces the classifier weight for class $i$ with its $c_{i}$ ($w_{i}=c_{i}$)
 and uses a cosine classifier for classification, i.e.:
 $$f(x)=\frac{W^{\top}\phi(x)}{||W||_{2}||\phi(x)||_{2}}$$
 Therefore, when faced with a new task, it is possible to calculate and replace each class's classifier with the embeddings frozen. This simple solution demonstrates superior performance compared to many prompt-based methods \cite{zhou2024expandable,zhou2024revisitingclassincrementallearningpretrained}. This indicates that PTMs already possess generalized representations that can be directly applied to downstream tasks. Similar phenomena have also been observed in large language models by \cite{janson2022simple} and \cite{zheng2023learn}. In our experiments, we will use this baseline to compare against state-of-the-art methods.
 
\section{Weight Balanced Replay}

\subsection{From classifier bias to network bias}

An intuitive approach to quickly align the parameter distributions of a forgetting model and a normal model is to identify and correct the biases between them. Thus, the focus lies in uncovering the distribution patterns of these biases. Earlier studies have identified traces of such patterns in the embedding module and classifier. For example, the embedding function outputs in a forgetting model are more concentrated \cite{shi2022mimicking}, the logits of new tasks are significantly higher than those of old tasks, and most importantly, the weight of new tasks in the classifier is greater than that of old tasks, which can lead to other forgetting characteristics. Furthermore, we hypothesize that all parameters in a forgetting model may follow this pattern. Ideally, we aim to discover a forgetting pattern for parameters that applies to any model optimized using backpropagation and gradient descent. Therefore, we define the continual learning process at stage $b$ as $f_{D_{b}}^{b-1}(x;\theta)$, where $f^{b-1}$ represents the model after completing the previous stage of continual learning, and the corresponding normal training process is denoted as $f_{D_{1} \cup \dots \cup D_{b}}(x;\theta)$, which involves training on data from $D_{1}$ to $D_{b}$. The goal of overcoming forgetting can thus be defined as:
\begin{equation}
    f_{D_{1} \cup \dots \cup D_{b}}(x;\theta) \gets f_{D_{b}}^{b-1}(x;\theta),
\end{equation}
Specifically, let $w^{b}$ denote the weights of $f^{b-1}$ after supervised learning on $D_{b}$, where these weights may exhibit forgetting. Correspondingly, let $w^{o}$ represent the weights obtained from $f_{D_{1} \cup \dots \cup D_{b}}$ trained without forgetting. The bias between them can then be defined as:
\begin{equation}
    w^{o} \gets w^{b}+\Delta w^{b},
\end{equation}
Previous work has focused on linearly adjusting classifier parameters to mitigate forgetting. For instance, some approaches use a cosine classifier to avoid biases in classifier weights \cite{hou2019learning}, apply weight normalization after training \cite{zhao2020maintaining}, or use additional parameters to scale classifier weights during training \cite{wu2019large}. In contrast, our approach aims to swiftly correct the biases caused by forgetting. It is important to note that we omit the bias term in the notation, as the impact of the weights on catastrophic forgetting is sufficiently significant.

\subsection{Approximate bias and balance}

We designed a replay-based approximate correction strategy to dynamically adjust the parameters during new task training (see Figure \ref{fig2}). This correction mechanism shares some design principles with sample buffering methods, which typically maintain a portion of original samples in memory to combat forgetting. However, our approach aims to obtain prior knowledge at minimal cost to quickly compute the bias $\Delta w^{b}$. We associate $\Delta w^{b}$ with each previously learned task. If this bias is distributed across each training iteration in the $b$-th stage of continual learning, considering the additivity of gradients, it can be easily expanded as follows:
\begin{equation}
    \Delta w^{b} =  \sum_{i=1}^{N_{1}}a_{i}^{1}\delta_{i}^{1 }+\dots+\sum_{i=1}^{N_{b-1}}a_{i}^{b-1}\delta_{i}^{b-1 },
\end{equation}
Here, $a$ represents the activation from the previous layer, and $\delta$ is the error backpropagated from the next layer. The product of these two gives the gradient update matrix for $w$. $\sum_{i=1}^{N_{b-1}}a_{i}^{b-1}\delta_{i}^{b-1 }$ denotes the actual bias of task $b-1$ in $\Delta w$. Ideally, the supervised learning process can be viewed as the gradient update generated by the samples of task $b$, plus the bias from tasks $1$ through $b-1$. In other words, as long as the cost of calculating the bias is sufficiently low, supervised learning can closely approximate continual learning. Our idea is that if the balance of weights between different tasks is the key factor influencing catastrophic forgetting, then retaining only the most essential information for weight updates should also lead to significant changes in forgetting, without the need to compute the forward and backward passes for all samples. To this end, we introduce an approximate strategy, similar to parameter regularization methods \cite{kirkpatrick2017overcoming}, to compute the importance of each parameter. We calculate the importance of each $a$:
\begin{equation}
    a_{*}=\sum_{i=1}^{N_{b}} \Omega (a_{i})*a_{i},
\end{equation}
$\Omega$ is an importance function that evaluates the significance of the current sample. Based on this, we can quickly calculate the bias:
\begin{equation}
    \Delta w^{b} \approx \Delta w^{b}_{*} = (a_{*}^{1}\delta_{*}^{1})+\dots+ (a_{*}^{b-1}\delta_{*}^{b-1}),
\end{equation}
While reducing the computational cost of bias, it does not resolve the balance issue during training. To address this, we use a simple yet effective solution that explicitly controls the update step sizes on both sides. Specifically, we apply different gradient constraints to the training of the current task and the calculation of bias:
\begin{equation}
    w^{o}\gets w^{b-1}+\min (\alpha ,\nabla w^{b})+\min (\beta ,\Delta w^{b}_{*}),
\end{equation}
$\nabla w^{b}$ represents the update gradient generated by supervised learning of $w^{b-1}$ on $D_{b}$, constrained by $\alpha$. Similarly, when calculating the bias for previous tasks, the process is constrained by $\beta$.

\textbf{Importance function. }Regarding the importance function, we found that average sampling already performs very well, similar to class centers:
\begin{equation}
    a_{*}= \frac{1}{N_{b}} \sum_{i=1}^{N_{b}}a_{i},
\end{equation}
We experimented with using confidence-weighted sampling similar to the herding strategy \cite{rebuffi2017icarl}:
\begin{equation}
    a_{*}= \sum_{i=1}^{N_{b}} (1-S_{b}(f(x_{i};\theta)))*a_{i},
\end{equation}
In conclusion, the latter only improved by 1\% over the former, making the improvement seem negligible and not particularly meaningful.

\begin{algorithm}[tb]
\caption{Weight Balanced Replay on PTM}
\label{alg:wbr}
\textbf{Input}: Pre-trained encoder $g$; training dataset $D_{b}$ for
task $b = 1,\dots, B$; decoder $f(g(x);\theta)$; learning rates $\gamma$ for $\theta$; clip factor $\alpha,\beta$.\\
\textbf{Initialization}: freeze encoder $g$; initialize $\theta$ randomly.\\
\begin{algorithmic}[1] 
\FOR{task $b = 1,\dots, B$}
\WHILE{$batch \gets D_{b}$}
\STATE Train $f(g(x);\theta)$ on $batch$. Option: $Clip(\theta,\alpha)$.
\STATE Train $f(g(x);\theta)$ on $memory$. Option: $Clip(\theta,\beta)$.
\ENDWHILE
\STATE $g(x)_{*}^{b}=\sum_{i=1}^{N_{b}} \Omega (x_{i})*g(x_{i})$
\STATE $memory=[g(x)_{*}^{1},\dots,g(x)_{*}^{b-1}]\gets g(x)_{*}^{b}$
\ENDFOR
\end{algorithmic}
\end{algorithm}

\subsection{Optimization objective for WBR}

Following the above strategy, the memory information $a_{*}$ for each task only needs to be calculated once and can be easily completed during its respective continual learning training phase, as training typically involves at least one pass through the data. In each training step, the data $x$ from the current task is input into the model alongside the memory information $x_{*}$ for supervised learning. While seeking weight balance, our objective is to minimize the end-to-end training loss function:
\begin{equation}
\label{loss}
\begin{aligned}
    \underset{\theta }{ \mathrm{ min}}\qquad   \mathcal{ L}(f(x;\theta ),y)+\sum_{1}^{b-1} \mathcal{ L}(f(x_{*}^{j};\theta ),y),\\
    s.t.\qquad Clip(f(x;\theta ),\alpha ),Clip(f(x_{*};\theta ),\beta  ),
\end{aligned}
\end{equation}
Both losses are computed using softmax cross-entropy and are controlled by the gradient clipping function to limit the gradient magnitude. If a pretrained model 
$g(x)$ is used, the loss function becomes:
\begin{equation}
\begin{aligned}
    \underset{\theta }{ \mathrm{ min}}\qquad  \mathcal{L}(f(g(x);\theta ),y)+\sum_{1}^{b-1} \mathcal{ L}(f(g(x)_{*}^{j};\theta ),y),\\
    s.t.\qquad Clip(f(g(x);\theta ),\alpha ),Clip(f(g(x)_{*};\theta ),\beta  ),
\end{aligned}
\end{equation}
$g(x)_{*}$ indicates that $g(x)$ is used in place of $x$ to compute the memory information.

\section{Experiments}

To evaluate the proposed WBR method, we strictly followed the experimental setup of previous works \cite{rebuffi2017icarl,zhou2024expandable} and conducted experiments in a class-incremental learning setting, where the task identity is unknown during inference. Specifically: (1)  We conducted ablation experiments in a simplified environment to more intuitively understand the relationship between this balance and catastrophic forgetting. (2) We compared our method with various state-of-the-art approaches on the latest pretrained model-based continual learning baseline \cite{zhou2024revisitingclassincrementallearningpretrained}. Finally, we discussed the results of WBR compared to other methods, as well as its implications for addressing catastrophic forgetting.

\begin{figure*}[t]
\centering
\begin{minipage}[t]{0.33\textwidth}
\centering
\includegraphics[width=5.8cm]{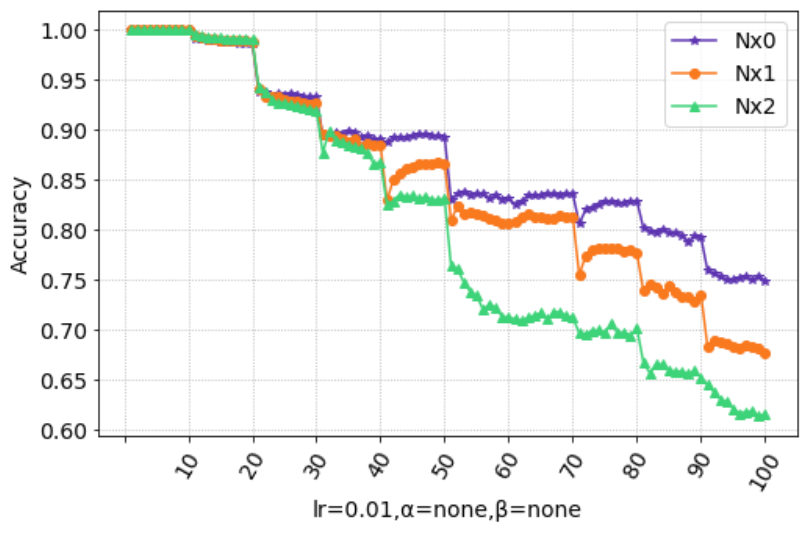}
\end{minipage}
\begin{minipage}[t]{0.33\textwidth}
\centering
\includegraphics[width=5.8cm]{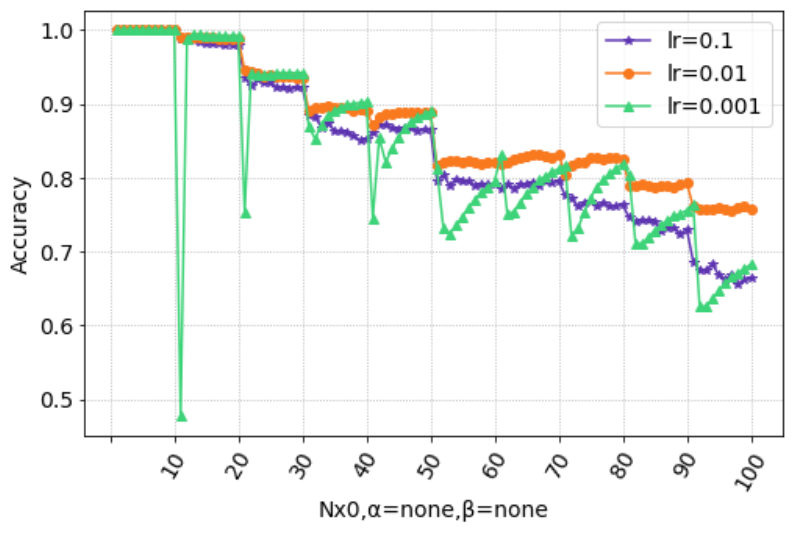}
\end{minipage}
\centering
\begin{minipage}[t]{0.33\textwidth}
\centering
\includegraphics[width=5.8cm]{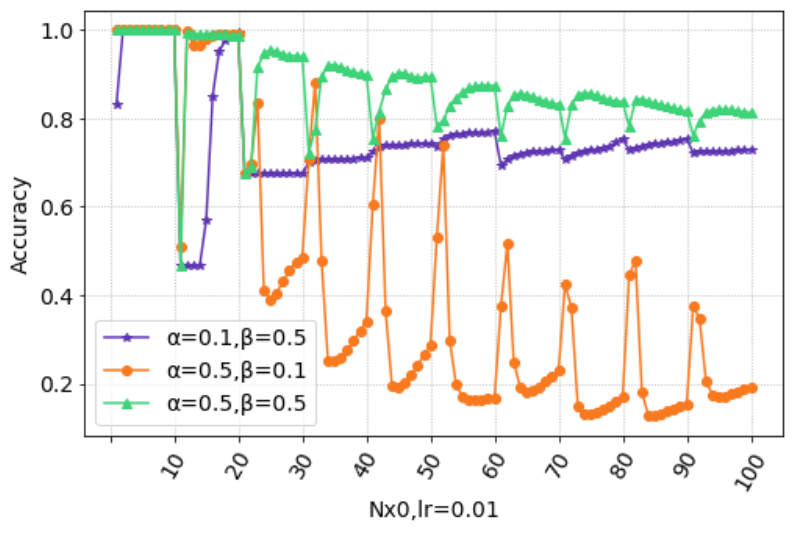}
\end{minipage}
\caption{Ablation Study on split MNIST. The left figure shows the impact of network depth on the control of the first layer's memory vector when $\alpha$ and $\beta$ are not set. The middle figure illustrates the effect of different learning rates on mitigating forgetting. The right figure depicts the impact of different learning constraints $\alpha$ and memory constraints $\beta$ on forgetting when the learning rate is set to 0.01 and no hidden layers are used.}
\label{fig:non-pre}
\end{figure*}

\subsection{Comparing methods}

We compared the WBR method with multiple baseline and state-of-the-art continual learning methods. Our approach is based on the pretrained ViT-B-16 model \cite{dosovitskiy2020image}, which has become a common tool in the field of pretrained model-based continual learning. To ensure a fair comparison, we utilized the latest pretrained continual learning baseline, SimpleCIL , which allows us to evaluate algorithm performance across different pretrained models. We referred to recent review papers \cite{zhou2024continual,zhou2023deep} and the latest research works, selecting the most recognized or best-performing methods in each domain.

\textbf{Baseline Methods}. Fine-tuning involves sequential supervised learning on all continual learning data and is typically regarded as a scenario where catastrophic forgetting occurs completely, serving as a classical baseline for continual learning. EWC \cite{kirkpatrick2017overcoming} is a representative method based on parameter regularization. SimpleCIL is a baseline method for pretrained model-based continual learning, representing the performance that can be achieved through the inherent generalization capabilities of pretrained models without learning any additional parameters, including the classifier.

\textbf{SOTA Sample Buffer-Based Methods}. We selected three state-of-the-art sample buffer-based methods for comparison, including iCaRL \cite{rebuffi2017icarl}, BiC \cite{wu2019large}, and WA \cite{zhao2020maintaining}. iCaRL is a representative method that improves upon sample buffers using knowledge distillation, while WA is a representative method that further rectify the model based on distillation.

\textbf{SOTA Architecture-Based Methods}. We selected two architecture-based methods that do not use pretrained models for comparison: DER \cite{yan2021dynamically} and MEMO \cite{zhou2022model}. These methods neither use sample buffers nor rely on pretrained models, making them typical examples of architecture expansion approaches.

\textbf{SOTA Pretrained Model-Based Methods}. We selected two state-of-the-art methods that use frozen pretrained models for comparison: ADAM \cite{zhou2024revisitingclassincrementallearningpretrained} and EASE \cite{zhou2024expandable}. Both methods utilize additional adapters for fine-tuning.

\textbf{Our Method}. WBR is the method we propose. Unlike other approaches, WBR does not use a sample buffer but instead maintains a minimal, non-learnable memory pool. Furthermore, WBR introduces no additional parameters and does not require any post-processing.

\begin{table}[t]
\centering
\begin{tabular}{l|l|l}
\toprule
\multirow{2}{*}{\textbf{Setting}}  & \multicolumn{2}{c}{Base-0,Inc-1} \\
& $A_{B}(\%)\uparrow $ &$\bar{A}(\%)\uparrow$\\
    \midrule
  $lr=0.01,N\times0$  & 74.87\tiny $\pm$0.28 & 87.4$\pm$\tiny 0.33  \\
  $lr=0.01,N\times1$  & 67.7 \tiny $\pm$0.15& 84.73 \tiny $\pm$0.1\\
  $lr=0.01,N\times2$  & 61.56 \tiny $\pm$0.47& 80.03  \tiny $\pm$0.13\\
  $lr=0.1,N\times0$  & 66.40 \tiny $\pm$0.62& 83.65 \tiny $\pm$0.6 \\
  $lr=0.01,N\times0$  & 75.69 \tiny $\pm$0.81& 87.30 \tiny $\pm$0.69\\
  $lr=0.001,N\times0$  & 68.29 \tiny $\pm$0.88& 85.91 \tiny $\pm$0.66 \\
  $lr=0.1,N\times2$  & 37.59 \tiny $\pm$0.51& 70.51 \tiny $\pm$0.14 \\
  $lr=0.01,N\times0,\alpha=0.5$  & \textbf{81.18 \tiny $\pm$0.17} & \textbf{89.18 \tiny $\pm$0.32}\\
\bottomrule
\end{tabular}
\caption{Ablation Study on split MNIST. The impact of different hyperparameter settings on forgetting.}
\label{table:non-pre}
\end{table}

\subsection{Datasets and experimental details}

\textbf{Datasets}. We used Split CIFAR-100 \cite{krizhevsky2009learning} in a class-incremental setting to evaluate the effectiveness of our method. Additionally, we conducted ablation studies on MNIST to assess the impact of network depth, learning rate, and the balance between $\alpha$ and $\beta$ on WBR's performance.

\textbf{Evaluation Metrics}. We used two evaluation metrics: final accuracy $A_{B}$ and average accuracy $\bar{A}$, both of which are better when higher. These metrics have been widely used in previous research \cite{zhao2020maintaining,zhou2024expandable,rebuffi2017icarl}. Specifically, $A_{b}$ represents the top-1 accuracy of the model across all learned tasks after the $b$-th stage of continual learning. $\bar{A}$ is calculated as $\frac{1}{B}\sum_{b=1}^{B}A_{b}$, which averages the accuracy at each stage.

\begin{table*}[t]
\centering
\begin{tabular}{l|l|l|l|l|l|l|l}
\toprule
\multirow{2}{*}{\textbf{Method}}  &
  \multicolumn{2}{c}{Base0,Inc-5}&\multicolumn{2}{c}{Base0,Inc-10} & \multirow{2}{*}{Epochs$\downarrow$}  &\multirow{2}{*}{Buffer size$\downarrow$}&\multirow{2}{*}{Extra param}\\
  &Last $A_{B}\uparrow $ &Average $\bar{A}\uparrow$ 
  &Last $A_{B}\uparrow $ &Average $\bar{A}\uparrow$\\
\midrule
    Fintune & 4.83 & 17.59 & 9.09 & 26.25 & 170 & - & n\\
    EWC & 5.58 & 18.42 & 12.44 & 29.73 & 170 & - & n\\
    \midrule
    iCaRL & 45.12 & 63.51 & 49.52 &  64.42 & 170 & 2000 & n\\
    WA & 48.46 & 64.65 & 52.30 & 67.09 & 170 & 2000 & n\\
    BiC & 43.08 & 62.38 & 50.79 & 65.08 & 170 & 2000 & y\\
    DER & 53.95 & 67.99 & \textbf{58.59} & 69.74 & 170 & - & y\\
    MEMO & \textbf{54.23} & \textbf{68.10} & 58.49 & \textbf{70.20} & 170 & - & y\\
    \midrule
    SimpleCIL & 64.41 & 76.1 & 64.39 & 75.46 & - & - & n\\
    \midrule
    ADAM + Finetune & +0.01 & +0.1 & - & - & 20 & - & n\\
    ADAM + VPT-S & +3.31 & +2.86 & - & - & 20 & - & y\\
    ADAM + VPT-D & +0.91 & +0.89 & - & - & 20 & - & y\\
    ADAM + SSF & +0.72 & +0.21 & - & - & 20 & - & y\\
    ADAM + Adapter & +3.89 & +3.08 & - & - & 20 & - & y\\
    EASE & \textbf{+4.54} & \textbf{+3.94} & - & - & 20 & - & y\\
    \midrule
    WBR,$\alpha=0.5$ & +3.91 & +2.21 & +4.37 & +2.79 & \textbf{3} & 23.75 & n\\
    WBR,$\alpha$ not set & -5.29 & -3.9 & -5.59 & -1.88 & \textbf{3} & 23.75 & n\\
    WBR,$moment=0.9$& -35.8 & -38.17 & -31.25 & -27.35 & \textbf{3} & 23.75 & n\\
    \bottomrule
\end{tabular}
\caption{Comparison Results on split CIFAR-100 for Class-Incremental Learning (Task Identity Unknown During Testing). In methods using pretrained models, we conducted comparisons based on the SimpleCIL benchmark, which utilizes models pretrained on the ImageNet1K dataset instead of ImageNet21K. WBR significantly outperforms other methods in terms of training speed, while achieving competitive results.}
\label{tab:pre}
\end{table*}

\textbf{Training Details}. For the WBR method, we used PyTorch and ran the experiments on an NVIDIA 3060 GPU. All experiments were conducted using the SGD optimizer for training with the following setups: \textit{Non-Pretrained Environment}. We used a Multilayer Perceptron (MLP) as the training model with the structure 784x(32xN)x10, where N represents the number of hidden layers. We experimented with three learning rates (0.1, 0.01, 0.001) and provided three options for the balance factors $\alpha$ and $\beta$: 0.1, 0.5, and not set. MNIST input images were resized to 784x1 to match the network structure, and each task was trained for 10 epochs. \textit{Pretrained Environment}. We primarily used ViT-B-16 as the encoder, with pretrained weights from the publicly available ImageNet1K weights in the Torchvision library. We only trained the classifier's parameters, while the encoder was frozen. Each task was trained for 3 epochs, with learning rates of 0.01, batch size set to 16. Since the training gradient for new tasks is usually higher than the gradient when calculating bias, we set $\alpha=0.5$ here and do not set $\beta$. Input images were resized to 224x224 and normalized to the [0,1] range to match the pretrained setup. Specifically, The output dimension of ViT-B-16 is 768, which defines the size of each memory vector. For instance, in the base-0, inc-5 setup, 95 such vectors need to be stored. For CIFAR-100, with each sample size being 32x32x3, this equals 23.75 sample sizes. Similarly, in the base-0, inc-10 setup, 90 vectors need to be stored, equaling 22.5, respectively. For Other Methods. According to the papers \cite{zhou2023deep,zhou2024revisitingclassincrementallearningpretrained,zhou2024expandable}, non-pretrained model methods use ResNet-32 as the backbone network, with the training period uniformly set to 170 epochs and a sample buffer size of 2000. Pretrained model methods use the ViT-B-16 model pretrained on ImageNet21K, with the training period uniformly set to 20 epochs. All methods utilize the SGD optimizer. Since the pretrained model weights differ from those in our experiments, we use SimpleCIL as the baseline for comparison.

\subsection{Main results}

\textbf{Results in the Non-Pretrained Environment.} Figure \ref{fig:non-pre} and Table \ref{table:non-pre} illustrates the performance of WBR in a non-pretrained environment. Overall, the results indicate that significant forgetting can be mitigated without the need for extensive mixed supervision training; maintaining a balance between the weights of old task and the new task is sufficient. Specifically: (1) It can be observed that as the depth of the network increases, the impact of this balance on forgetting significantly diminishes. (2) The learning rate also has a notable effect on this balance: too high a learning rate exacerbates forgetting, while too low a learning rate hinders the timely acquisition of new knowledge. An appropriate learning rate keeps learning and forgetting in a relatively stable state. (3) Catastrophic forgetting is highly correlated with this balance. Using the analogy of stability versus plasticity is apt: when $\alpha$ is greater than $\beta$, classical catastrophic forgetting occurs, meaning plasticity outweighs stability; when $\alpha$ is less than $\beta$, learning efficiency is low, meaning stability outweighs plasticity. When $\alpha$ equals $\beta$, a balance is achieved, effectively controlling forgetting. However, we can also observe that even without applying balance constraints, simply incorporating memory vectors into the training process has already mitigated forgetting to some extent. Additionally, since the gradients of new tasks are typically higher than those of the memory vectors during actual training, adjusting the gradient constraints for new tasks alone is more efficient.

\begin{figure*}[t]
\centering
\begin{minipage}[t]{0.24\textwidth}
\centering
\includegraphics[width=4.4cm]{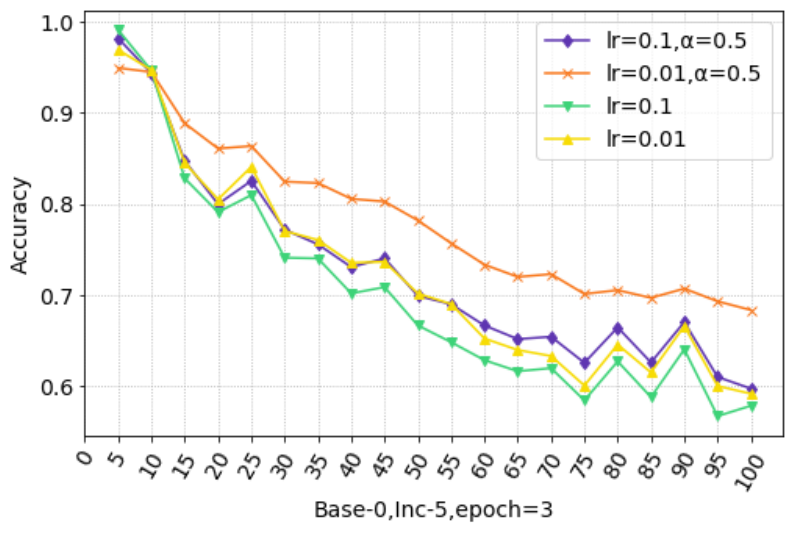}
\end{minipage}
\begin{minipage}[t]{0.24\textwidth}
\centering
\includegraphics[width=4.4cm]{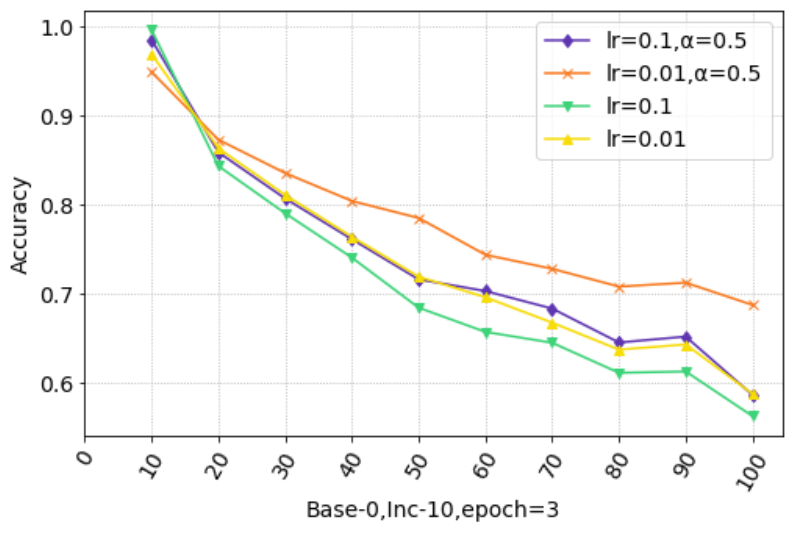}
\end{minipage}
\centering
\begin{minipage}[t]{0.24\textwidth}
\centering
\includegraphics[width=4.4cm]{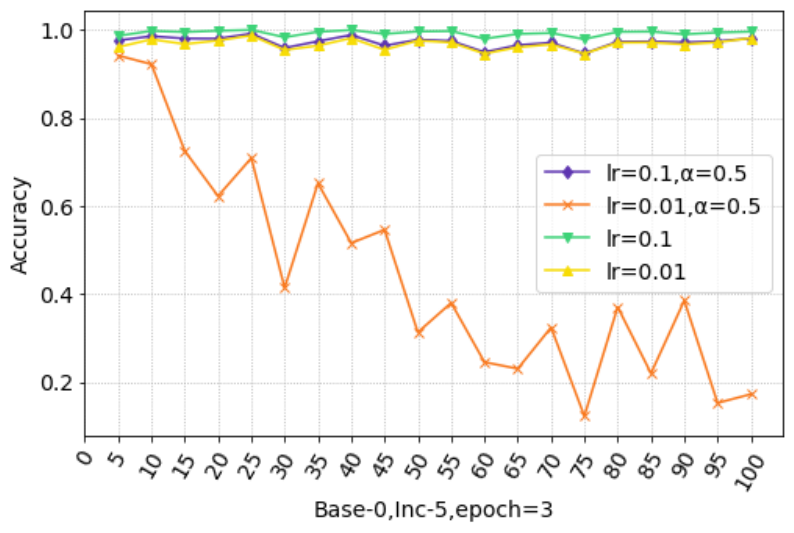}
\end{minipage}
\centering
\begin{minipage}[t]{0.24\textwidth}
\centering
\includegraphics[width=4.4cm]{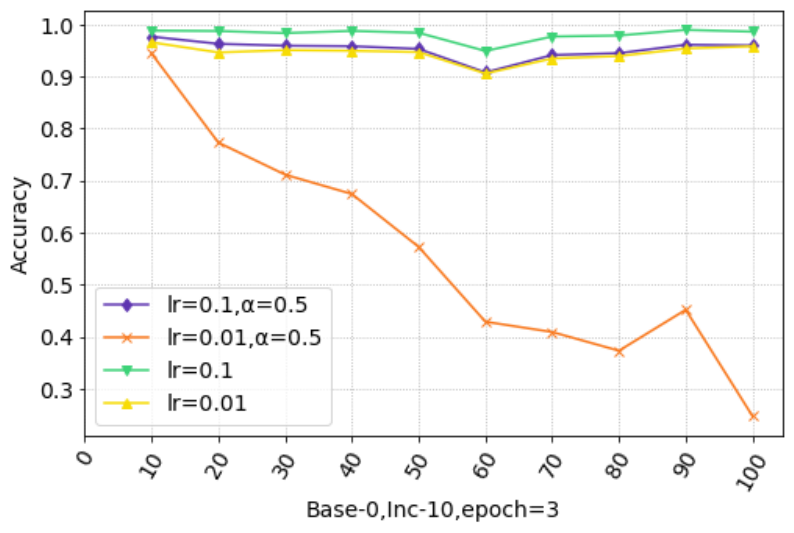}
\end{minipage}
\caption{The two figures on the left show the results of WBR based on ViT-B-16 (pretrained on ImageNet1K) on Split CIFAR-100, while the right figure displays the accuracy of learning new tasks at each stage. Notably, when the hyperparameters are appropriately set, the accuracy of new tasks ceases to improve and instead decreases. Catastrophic forgetting typically occurs when new tasks overfit, causing the performance on old tasks to degrade. This suggests that, at least locally, the balance of weights is closely related to catastrophic forgetting.}
\label{fig:pre}
\end{figure*}

\textbf{Results in the Pretrained Environment}. Table \ref{tab:pre} summarizes the results on the class-incremental benchmarks. WBR outperforms all comparison methods in terms of training speed and is competitive with state-of-the-art methods in performance. We attribute this to WBR's use of prior knowledge to directly mitigate catastrophic forgetting, rather than relying on indirect means. The results in Figure \ref{fig:pre} further support this: when the weights are well-balanced, the performance of new tasks cannot continue to improve, indicating that the model may have reached the limits of the pretrained model's representation capacity. In contrast, other settings show near 100\% accuracy, which clearly indicates overfitting and the occurrence of forgetting. Additionally, when momentum is enabled, the performance of WBR significantly decreases, which we believe leads to a disruption of balance. This also aligns with our design of the bias, as we did not account for the changes brought by momentum, but rather focused solely on the updates driven by the original gradients.

\section{Limitations}

As shown in Figure \ref{fig:non-pre}, with increasing network depth, the influence of the memory vector from the first layer on the subsequent layers weakens, which contradicts the advantages of deep learning. The power of deep learning comes from the nonlinear stacking of multiple layers, which results in strong fitting capabilities, something that a single-layer network lacks. This is precisely why WBR is well-suited for pretrained models; it only needs to address forgetting locally, without dealing with the diminishing influence in deeper networks. 

Furthermore, using only MLPs for experiments is, strictly speaking, incomplete. We have demonstrated the correlation between catastrophic forgetting and this weight balance from the perspective of backpropagation and gradient descent. However, in more complex architectures, such as convolutional or attention modules, we do not know how to design this balance. More critically, if we were to store memory vectors for each layer of the network, a significant challenge would be that after backpropagation updates, the memory vectors for deeper layers could become invalid due to changes in the earlier layers. It is foreseeable that attempting to address these challenges in practice would involve significant complexity and engineering effort. Perhaps, instead of designing more intricate training processes, it might be better to explore an algorithm different from backpropagation.

\section{Conclusion}

This paper proposes a novel method to overcome catastrophic forgetting, aiming to demonstrate that imbalances in any weights, not just in the classifier, are one of the direct causes of catastrophic forgetting. WBR extends the concept of classifier bias to arbitrary weights and introduces the use of approximate information instead of sample information to compute weight biases, while controlling weight balance during training through gradient constraints. This approach effectively and rapidly corrects the weight biases caused by catastrophic forgetting. The method significantly outperforms previous state-of-the-art approaches in training speed for class-incremental learning without sacrificing performance.

\bibliography{aaai25}

\begin{thebibliography}{40}
\providecommand{\natexlab}[1]{#1}

\bibitem[{Beaulieu et~al.(2020)Beaulieu, Frati, Miconi, Lehman, Stanley, Clune, and Cheney}]{beaulieu2020learning}
Beaulieu, S.; Frati, L.; Miconi, T.; Lehman, J.; Stanley, K.~O.; Clune, J.; and Cheney, N. 2020.
\newblock Learning to continually learn.
\newblock In \emph{ECAI 2020}, 992--1001. IOS Press.

\bibitem[{Buzzega et~al.(2020)Buzzega, Boschini, Porrello, Abati, and Calderara}]{buzzega2020dark}
Buzzega, P.; Boschini, M.; Porrello, A.; Abati, D.; and Calderara, S. 2020.
\newblock Dark experience for general continual learning: a strong, simple baseline.
\newblock \emph{Advances in neural information processing systems}, 33: 15920--15930.

\bibitem[{Cao et~al.(2023)Cao, Tang, Lin, Jiang, Dong, Han, Chen, Wang, and Sun}]{cao2023retentive}
Cao, B.; Tang, Q.; Lin, H.; Jiang, S.; Dong, B.; Han, X.; Chen, J.; Wang, T.; and Sun, L. 2023.
\newblock Retentive or forgetful? diving into the knowledge memorizing mechanism of language models.
\newblock \emph{arXiv preprint arXiv:2305.09144}.

\bibitem[{Castro et~al.(2018)Castro, Mar{\'\i}n-Jim{\'e}nez, Guil, Schmid, and Alahari}]{castro2018end}
Castro, F.~M.; Mar{\'\i}n-Jim{\'e}nez, M.~J.; Guil, N.; Schmid, C.; and Alahari, K. 2018.
\newblock End-to-end incremental learning.
\newblock In \emph{Proceedings of the European conference on computer vision (ECCV)}, 233--248.

\bibitem[{Cha, Lee, and Shin(2021)}]{cha2021co2l}
Cha, H.; Lee, J.; and Shin, J. 2021.
\newblock Co2l: Contrastive continual learning.
\newblock In \emph{Proceedings of the IEEE/CVF International conference on computer vision}, 9516--9525.

\bibitem[{Chamikara et~al.(2018)Chamikara, Bert{\'o}k, Liu, Camtepe, and Khalil}]{chamikara2018efficient}
Chamikara, M. A.~P.; Bert{\'o}k, P.; Liu, D.; Camtepe, S.; and Khalil, I. 2018.
\newblock Efficient data perturbation for privacy preserving and accurate data stream mining.
\newblock \emph{Pervasive and Mobile Computing}, 48: 1--19.

\bibitem[{Chaudhry et~al.(2018{\natexlab{a}})Chaudhry, Dokania, Ajanthan, and Torr}]{chaudhry2018riemannian}
Chaudhry, A.; Dokania, P.~K.; Ajanthan, T.; and Torr, P.~H. 2018{\natexlab{a}}.
\newblock Riemannian walk for incremental learning: Understanding forgetting and intransigence.
\newblock In \emph{Proceedings of the European conference on computer vision (ECCV)}, 532--547.

\bibitem[{Chaudhry et~al.(2018{\natexlab{b}})Chaudhry, Ranzato, Rohrbach, and Elhoseiny}]{chaudhry2018efficient}
Chaudhry, A.; Ranzato, M.; Rohrbach, M.; and Elhoseiny, M. 2018{\natexlab{b}}.
\newblock Efficient lifelong learning with a-gem.
\newblock \emph{arXiv preprint arXiv:1812.00420}.

\bibitem[{Chen et~al.(2022)Chen, Ge, Tong, Wang, Song, Wang, and Luo}]{chen2022adaptformer}
Chen, S.; Ge, C.; Tong, Z.; Wang, J.; Song, Y.; Wang, J.; and Luo, P. 2022.
\newblock Adaptformer: Adapting vision transformers for scalable visual recognition.
\newblock \emph{Advances in Neural Information Processing Systems}, 35: 16664--16678.

\bibitem[{De~Lange et~al.(2021)De~Lange, Aljundi, Masana, Parisot, Jia, Leonardis, Slabaugh, and Tuytelaars}]{de2021continual}
De~Lange, M.; Aljundi, R.; Masana, M.; Parisot, S.; Jia, X.; Leonardis, A.; Slabaugh, G.; and Tuytelaars, T. 2021.
\newblock A continual learning survey: Defying forgetting in classification tasks.
\newblock \emph{IEEE transactions on pattern analysis and machine intelligence}, 44(7): 3366--3385.

\bibitem[{Dosovitskiy et~al.(2020)Dosovitskiy, Beyer, Kolesnikov, Weissenborn, Zhai, Unterthiner, Dehghani, Minderer, Heigold, Gelly et~al.}]{dosovitskiy2020image}
Dosovitskiy, A.; Beyer, L.; Kolesnikov, A.; Weissenborn, D.; Zhai, X.; Unterthiner, T.; Dehghani, M.; Minderer, M.; Heigold, G.; Gelly, S.; et~al. 2020.
\newblock An image is worth 16x16 words: Transformers for image recognition at scale.
\newblock \emph{arXiv preprint arXiv:2010.11929}.

\bibitem[{Gao et~al.(2023)Gao, Zhao, Sun, Xi, Zhang, Ghanem, and Zhang}]{gao2023unified}
Gao, Q.; Zhao, C.; Sun, Y.; Xi, T.; Zhang, G.; Ghanem, B.; and Zhang, J. 2023.
\newblock A unified continual learning framework with general parameter-efficient tuning.
\newblock In \emph{Proceedings of the IEEE/CVF International Conference on Computer Vision}, 11483--11493.

\bibitem[{Hou et~al.(2019)Hou, Pan, Loy, Wang, and Lin}]{hou2019learning}
Hou, S.; Pan, X.; Loy, C.~C.; Wang, Z.; and Lin, D. 2019.
\newblock Learning a unified classifier incrementally via rebalancing.
\newblock In \emph{Proceedings of the IEEE/CVF conference on computer vision and pattern recognition}, 831--839.

\bibitem[{Janson et~al.(2022)Janson, Zhang, Aljundi, and Elhoseiny}]{janson2022simple}
Janson, P.; Zhang, W.; Aljundi, R.; and Elhoseiny, M. 2022.
\newblock A simple baseline that questions the use of pretrained-models in continual learning.
\newblock \emph{arXiv preprint arXiv:2210.04428}.

\bibitem[{Jia et~al.(2022)Jia, Tang, Chen, Cardie, Belongie, Hariharan, and Lim}]{jia2022visual}
Jia, M.; Tang, L.; Chen, B.-C.; Cardie, C.; Belongie, S.; Hariharan, B.; and Lim, S.-N. 2022.
\newblock Visual prompt tuning.
\newblock In \emph{European Conference on Computer Vision}, 709--727. Springer.

\bibitem[{Ke, Liu, and Huang(2020)}]{ke2020continual}
Ke, Z.; Liu, B.; and Huang, X. 2020.
\newblock Continual learning of a mixed sequence of similar and dissimilar tasks.
\newblock \emph{Advances in neural information processing systems}, 33: 18493--18504.

\bibitem[{Kirkpatrick et~al.(2017)Kirkpatrick, Pascanu, Rabinowitz, Veness, Desjardins, Rusu, Milan, Quan, Ramalho, Grabska-Barwinska et~al.}]{kirkpatrick2017overcoming}
Kirkpatrick, J.; Pascanu, R.; Rabinowitz, N.; Veness, J.; Desjardins, G.; Rusu, A.~A.; Milan, K.; Quan, J.; Ramalho, T.; Grabska-Barwinska, A.; et~al. 2017.
\newblock Overcoming catastrophic forgetting in neural networks.
\newblock \emph{Proceedings of the national academy of sciences}, 114(13): 3521--3526.

\bibitem[{Krizhevsky, Hinton et~al.(2009)}]{krizhevsky2009learning}
Krizhevsky, A.; Hinton, G.; et~al. 2009.
\newblock Learning multiple layers of features from tiny images.

\bibitem[{Kumaran, Hassabis, and McClelland(2016)}]{kumaran2016learning}
Kumaran, D.; Hassabis, D.; and McClelland, J.~L. 2016.
\newblock What learning systems do intelligent agents need? Complementary learning systems theory updated.
\newblock \emph{Trends in cognitive sciences}, 20(7): 512--534.

\bibitem[{Mai et~al.(2022)Mai, Li, Jeong, Quispe, Kim, and Sanner}]{mai2022online}
Mai, Z.; Li, R.; Jeong, J.; Quispe, D.; Kim, H.; and Sanner, S. 2022.
\newblock Online continual learning in image classification: An empirical survey.
\newblock \emph{Neurocomputing}, 469: 28--51.

\bibitem[{McCloskey and Cohen(1989)}]{mccloskey1989catastrophic}
McCloskey, M.; and Cohen, N.~J. 1989.
\newblock Catastrophic interference in connectionist networks: The sequential learning problem.
\newblock In \emph{Psychology of learning and motivation}, volume~24, 109--165. Elsevier.

\bibitem[{Razdaibiedina et~al.(2023)Razdaibiedina, Mao, Hou, Khabsa, Lewis, and Almahairi}]{razdaibiedina2023progressive}
Razdaibiedina, A.; Mao, Y.; Hou, R.; Khabsa, M.; Lewis, M.; and Almahairi, A. 2023.
\newblock Progressive prompts: Continual learning for language models.
\newblock \emph{arXiv preprint arXiv:2301.12314}.

\bibitem[{Rebuffi et~al.(2017)Rebuffi, Kolesnikov, Sperl, and Lampert}]{rebuffi2017icarl}
Rebuffi, S.-A.; Kolesnikov, A.; Sperl, G.; and Lampert, C.~H. 2017.
\newblock icarl: Incremental classifier and representation learning.
\newblock In \emph{Proceedings of the IEEE conference on Computer Vision and Pattern Recognition}, 2001--2010.

\bibitem[{Shi et~al.(2022)Shi, Zhou, Liang, Jiang, Feng, Torr, Bai, and Tan}]{shi2022mimicking}
Shi, Y.; Zhou, K.; Liang, J.; Jiang, Z.; Feng, J.; Torr, P.~H.; Bai, S.; and Tan, V.~Y. 2022.
\newblock Mimicking the oracle: An initial phase decorrelation approach for class incremental learning.
\newblock In \emph{Proceedings of the IEEE/CVF conference on computer vision and pattern recognition}, 16722--16731.

\bibitem[{Shin et~al.(2017)Shin, Lee, Kim, and Kim}]{shin2017continual}
Shin, H.; Lee, J.~K.; Kim, J.; and Kim, J. 2017.
\newblock Continual learning with deep generative replay.
\newblock \emph{Advances in neural information processing systems}, 30.

\bibitem[{Smith et~al.(2023)Smith, Karlinsky, Gutta, Cascante-Bonilla, Kim, Arbelle, Panda, Feris, and Kira}]{smith2023coda}
Smith, J.~S.; Karlinsky, L.; Gutta, V.; Cascante-Bonilla, P.; Kim, D.; Arbelle, A.; Panda, R.; Feris, R.; and Kira, Z. 2023.
\newblock Coda-prompt: Continual decomposed attention-based prompting for rehearsal-free continual learning.
\newblock In \emph{Proceedings of the IEEE/CVF Conference on Computer Vision and Pattern Recognition}, 11909--11919.

\bibitem[{Wang et~al.(2022{\natexlab{a}})Wang, Zhou, Ye, and Zhan}]{wang2022foster}
Wang, F.-Y.; Zhou, D.-W.; Ye, H.-J.; and Zhan, D.-C. 2022{\natexlab{a}}.
\newblock Foster: Feature boosting and compression for class-incremental learning.
\newblock In \emph{European conference on computer vision}, 398--414. Springer.

\bibitem[{Wang et~al.(2023)Wang, Ma, Huang, Wang, Su, and Hong}]{wang2023isolation}
Wang, Y.; Ma, Z.; Huang, Z.; Wang, Y.; Su, Z.; and Hong, X. 2023.
\newblock Isolation and impartial aggregation: A paradigm of incremental learning without interference.
\newblock In \emph{Proceedings of the AAAI Conference on Artificial Intelligence}, volume~37, 10209--10217.

\bibitem[{Wang et~al.(2022{\natexlab{b}})Wang, Zhang, Lee, Zhang, Sun, Ren, Su, Perot, Dy, and Pfister}]{wang2022learning}
Wang, Z.; Zhang, Z.; Lee, C.-Y.; Zhang, H.; Sun, R.; Ren, X.; Su, G.; Perot, V.; Dy, J.; and Pfister, T. 2022{\natexlab{b}}.
\newblock Learning to prompt for continual learning.
\newblock In \emph{Proceedings of the IEEE/CVF conference on computer vision and pattern recognition}, 139--149.

\bibitem[{Wu et~al.(2019)Wu, Chen, Wang, Ye, Liu, Guo, and Fu}]{wu2019large}
Wu, Y.; Chen, Y.; Wang, L.; Ye, Y.; Liu, Z.; Guo, Y.; and Fu, Y. 2019.
\newblock Large scale incremental learning.
\newblock In \emph{Proceedings of the IEEE/CVF conference on computer vision and pattern recognition}, 374--382.

\bibitem[{Yan, Xie, and He(2021)}]{yan2021dynamically}
Yan, S.; Xie, J.; and He, X. 2021.
\newblock Der: Dynamically expandable representation for class incremental learning.
\newblock In \emph{Proceedings of the IEEE/CVF conference on computer vision and pattern recognition}, 3014--3023.

\bibitem[{Zenke, Poole, and Ganguli(2017)}]{zenke2017continual}
Zenke, F.; Poole, B.; and Ganguli, S. 2017.
\newblock Continual learning through synaptic intelligence.
\newblock In \emph{International conference on machine learning}, 3987--3995. PMLR.

\bibitem[{Zhang et~al.(2023)Zhang, Wang, Kang, Chen, and Wei}]{zhang2023slca}
Zhang, G.; Wang, L.; Kang, G.; Chen, L.; and Wei, Y. 2023.
\newblock Slca: Slow learner with classifier alignment for continual learning on a pre-trained model.
\newblock In \emph{Proceedings of the IEEE/CVF International Conference on Computer Vision}, 19148--19158.

\bibitem[{Zhao et~al.(2020)Zhao, Xiao, Gan, Zhang, and Xia}]{zhao2020maintaining}
Zhao, B.; Xiao, X.; Gan, G.; Zhang, B.; and Xia, S.-T. 2020.
\newblock Maintaining discrimination and fairness in class incremental learning.
\newblock In \emph{Proceedings of the IEEE/CVF conference on computer vision and pattern recognition}, 13208--13217.

\bibitem[{Zheng, Qiu, and Ma(2023)}]{zheng2023learn}
Zheng, J.; Qiu, S.; and Ma, Q. 2023.
\newblock Learn or Recall? Revisiting Incremental Learning with Pre-trained Language Models.
\newblock \emph{arXiv preprint arXiv:2312.07887}.

\bibitem[{Zhou et~al.(2024{\natexlab{a}})Zhou, Cai, Ye, Zhan, and Liu}]{zhou2024revisitingclassincrementallearningpretrained}
Zhou, D.-W.; Cai, Z.-W.; Ye, H.-J.; Zhan, D.-C.; and Liu, Z. 2024{\natexlab{a}}.
\newblock Revisiting Class-Incremental Learning with Pre-Trained Models: Generalizability and Adaptivity are All You Need.
\newblock arXiv:2303.07338.

\bibitem[{Zhou et~al.(2024{\natexlab{b}})Zhou, Sun, Ning, Ye, and Zhan}]{zhou2024continual}
Zhou, D.-W.; Sun, H.-L.; Ning, J.; Ye, H.-J.; and Zhan, D.-C. 2024{\natexlab{b}}.
\newblock Continual learning with pre-trained models: A survey.
\newblock \emph{arXiv preprint arXiv:2401.16386}.

\bibitem[{Zhou et~al.(2024{\natexlab{c}})Zhou, Sun, Ye, and Zhan}]{zhou2024expandable}
Zhou, D.-W.; Sun, H.-L.; Ye, H.-J.; and Zhan, D.-C. 2024{\natexlab{c}}.
\newblock Expandable subspace ensemble for pre-trained model-based class-incremental learning.
\newblock In \emph{Proceedings of the IEEE/CVF Conference on Computer Vision and Pattern Recognition}, 23554--23564.

\bibitem[{Zhou et~al.(2023)Zhou, Wang, Qi, Ye, Zhan, and Liu}]{zhou2023deep}
Zhou, D.-W.; Wang, Q.-W.; Qi, Z.-H.; Ye, H.-J.; Zhan, D.-C.; and Liu, Z. 2023.
\newblock Deep class-incremental learning: A survey.
\newblock \emph{arXiv preprint arXiv:2302.03648}.

\bibitem[{Zhou et~al.(2022)Zhou, Wang, Ye, and Zhan}]{zhou2022model}
Zhou, D.-W.; Wang, Q.-W.; Ye, H.-J.; and Zhan, D.-C. 2022.
\newblock A model or 603 exemplars: Towards memory-efficient class-incremental learning.
\newblock \emph{arXiv preprint arXiv:2205.13218}.

\end{thebibliography}

\end{document}